\documentclass[conference]{IEEEtran}
\IEEEoverridecommandlockouts
\usepackage{cite}
\usepackage{amsmath,amssymb,amsfonts}
\usepackage{algorithmic}
\usepackage{graphicx}
\usepackage{textcomp}
\usepackage{xcolor}
\usepackage{booktabs} 
\usepackage{subfigure}
\usepackage[linesnumbered]{algorithm2e}
\usepackage{bm}
\usepackage{multirow}
\usepackage{float}
\usepackage{stfloats}
\usepackage[T1]{fontenc}
\def\BibTeX{{\rm B\kern-.05em{\sc i\kern-.025em b}\kern-.08em
    T\kern-.1667em\lower.7ex\hbox{E}\kern-.125emX}}
\newcommand{\tabincell}[2]{
\begin{tabular}{@{}#1@{}}#2\end{tabular}
}    

\begin{document}
\title{AXNet: ApproXimate computing using an end-to-end trainable neural network
\thanks{This research was partially supported by National Natural Science Foundation of China (Grant No. 61602300), 
Shanghai Science and Technology Committee (Grant No. 18ZR1421400), 
Shanghai Jiao Tong University Biomedical Engineering Research Foundation (No. YG2015MS17), 
and Shanghai clinical ability construction of The three grade hospital (No. SHDC12015904). 
The Corresponding author is Li Jiang.}}

\author{\IEEEauthorblockN{Zhenghao Peng\IEEEauthorrefmark{1}, Xuyang Chen\IEEEauthorrefmark{1}, Chengwen Xu\IEEEauthorrefmark{1}, Naifeng Jing\IEEEauthorrefmark{1}, Xiaoyao Liang\IEEEauthorrefmark{1}, Cewu Lu\IEEEauthorrefmark{1}, Li Jiang\IEEEauthorrefmark{1}\IEEEauthorrefmark{2}}
\IEEEauthorblockA{\IEEEauthorrefmark{2}\textit{MoE Key Lab of Artificial Intelligence, AI Institute}\\
\IEEEauthorrefmark{1}\textit{School of Electronic, Information and Electrical Engineering, Shanghai Jiao Tong University}}}

\maketitle

\begin{abstract}
Neural network based approximate computing is a universal architecture promising to gain tremendous energy-efficiency for many error resilient applications. To guarantee the approximation quality, existing works deploy two neural networks (NNs), e.g., an approximator and a predictor. The approximator provides the approximate results, while the predictor predicts whether the input data is safe to approximate with the given quality requirement. However, it is non-trivial and time-consuming to make these two neural network coordinate---they have different optimization objectives---by training them separately. This paper proposes a novel neural network structure---AXNet---to fuse two NNs to a holistic end-to-end trainable NN. Leveraging the philosophy of multi-task learning, AXNet can tremendously improve the invocation (proportion of safe-to-approximate samples) and reduce the approximation error. The training effort also decrease significantly. Experiment results show 50.7\% more invocation and substantial cuts of training time when compared to existing neural network based approximate computing framework.
\end{abstract}

\renewcommand{\thefootnote}{}

\section{Introduction}
The conflict between increasing demand for computing and sluggish grow of hardware capability triggers the heated development of approximate computing, which has achieved massive success in both industry and research community.
Many applications that do not require utterly accurate computation can achieve tremendous acceleration and drastic reduction of the energy consumption by leveraging approximate computing, especially in domains that call for real-time calculation, fast response and low power consumption such as learning~\cite{zhang2015approxann:}, image processing~\cite{samadi2013sage:} and scientific computation~\cite{xu2015exploring}. 
Approximation computing can be conduct in different hierarchies, such as hardware \cite{ernst2003razor}, system and software levels. Various approximate computing architectures~\cite{zhang2015approxann:,samadi2013sage:,Mahajan2016Towards} are advocated. 

Neural network (NN) based approximate computing 
focus on the acceleration in software-level and has many advantages when compared to previous methods. First, neural networks are proved to be able to fit any continuous function~\cite{Hornik1991Approximation}, and thus this method can universally be adopted by different tasks.
Second, enormous parallelism in the neural networks is exploited by the rapid advancement of various neural network accelerators. An appropriate NN can be easily deserialized and deployed in the cloud~\cite{Chen2014DaDianNao} and on the edge~\cite{han2016eie} and therefore achieve high speedup. 

However,  a single neural network is not safe to serve as an accelerator due to the lack of approximation quality control.
Various metrics can represent the approximation quality, e.g., the mean-square error and absolute error between the approximated value and true value, etc.
Constraining those metrics can algorithm control the approximation quality. 
Dictinctive quality control mechanisms, such as statistical and linear models~\cite{Mahajan2016Towards}, Bayes-network~\cite{sui2016proactive}, neural network~\cite{Mahajan2015neuralprediction}, are proactively used to predict whether the approximator can safely approximate the output given the input data.
Those unsafe input data are sent to CPU for exact computation. 
On the contrary, predictors can also posteriorly monitor the output and determine the quality of approximation at the run-time~\cite{khudia2015rumba}. 
Predicted errors exceeding the error-bound incurs a rollback of execution~\cite{wang2016effective}. This architecture can dynamically adjust the approximator at run-time but takes more computation effort. Previous work reports that the neural network based predictor outperforms others regarding the prediction accuracy~\cite{Mahajan2016Towards}. 

New challenges emerge if both the approximate accelerator and predictor employ a neural network~\cite{Mahajan2015neuralprediction}, denoted as the \emph{approximator} and \emph{predictor} for simplicity, respectively. Mahajan et al.~\cite{Mahajan2016Towards} first train the best approximator and consequently the best predictor separately. The ignorance of the interaction between those two NNs plunges the approximate computing to a local optimum. To cope with this issue, Xu et al.~\cite{Xu2017Iterative} propose to iteratively and alternately train the approximator and predictor, by judicious selection of the training data in each iteration. This method reduces the approximation error. However, it inevitably causes exceptionally long training time. All these methods fail to find efficient cooperation of both NNs that produces the best speedup and approximation accuracy.

The obstacle to making the two NNs cooperation is that two NNs in the approximate computing framework---although share the same training data---have different tasks: prediction and regression. Inspired by multi-task learning~\cite{caruana1993multitask}, this paper presents a novel neural network structure, namely~\textbf{AXNet}. 
Instead of the weight sharing---a conventional method---AXNet fuses the approximator and predictor together, so that of AXNet a simple modification of the conventional back-propagation algorithm can train AXNet efficiently and effectively. We further propose a cost-effective deployment in a typical NPU design. 
To our best knowledge, AXNet is the first neural approximator that can adopt the end-to-end learning; \textbf{the proposed network fusion method has not seen in any previous work in machine learning domain.}

The rest of the paper is organized as follows. Section~\ref{sect:relatedwork} introduces the related works and motivation. Section~\ref{sect:approach} describes the proposed AXNet structure, the fusion methodology and the training algorithm. Section \ref{sect:architecture} shows a case study on the deployment of AXNet in a typical NPU. Experimental results are visualized and analyzed in section~\ref{sect:experiment}. Finally, section~\ref{sect:conclusion} concludes this paper. 
\section{Related Works and Motivation}\label{sect:relatedwork}
This section first introduces the related works on neural approximate computing frameworks containing the predictor and approximator and then motivates this paper.

\begin{figure}[bp]
\subfigure[]{
\centering
    \includegraphics[width=0.46\linewidth]{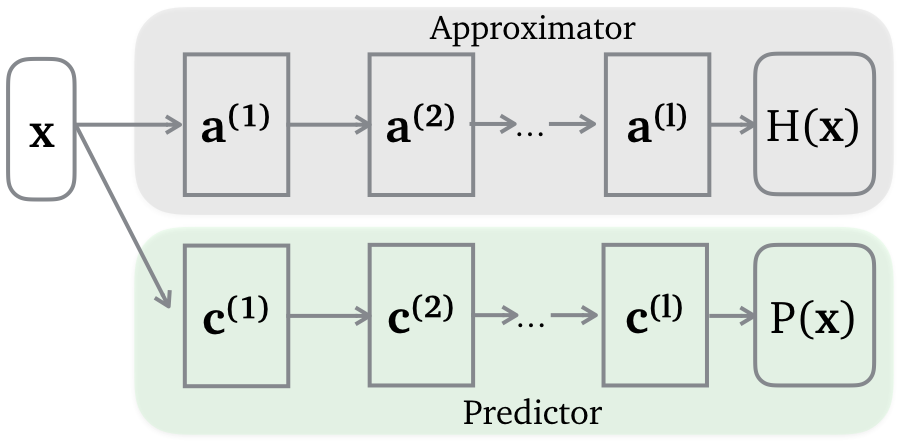}
    \label{fig:previous_network}}
    \subfigure[]{
    \includegraphics[width=0.46\linewidth]{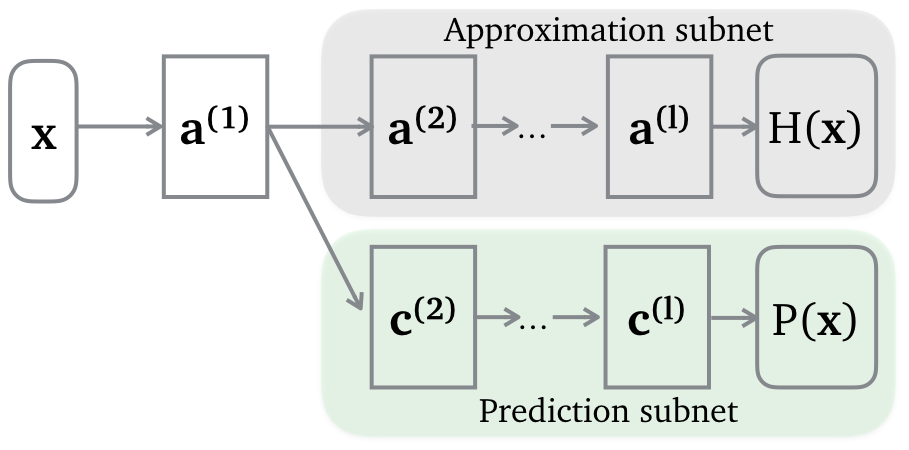}
    \label{fig:weight_sharing}}
    \caption{(a) Existing structure with standalone neuron networks. (b) Weight-sharing structure that we try.}
\end{figure}

Mahajan et al.~\cite{Mahajan2015neuralprediction} propose an approximate computing architecture consisting of a neural approximator and a neural predictor (Figure~\ref{fig:previous_network}). 
First, the approximator is trained to minimize its approximation error. In the training process, the input data of the target function entrances the approximator; the output of the approximator compares with the exact output value of the target function. The square-error between the approximate and exact output values is defined as the lost function. Then, they validate the approximator using the same set of the input data and derive a series of approximation results and consequently the approximation errors. The input data is labeled as \emph{safe-to-approximate} if the resulting approximation error is within the user-defined error-bound. Then, the predictor is trained using pairs of the input data and the derived label. In this method, the approximator and predictor are trained once, denoted as ``onepass'' training. In this neural approximate computing framework, however, only the "safe-to-approximate" data identified by the predictor can invoke the approximator. The effective approximation error only accounts those  ``safe-to-approximate'' data (Input data leading to significant approximation error will never enter the approximator). Thus, solely optimizing the approximator cannot efficiently minimize the approximation error of the whole framework. In onepass training method, the training process of the approximator and that of the predictor are isolated. There is no feedback from the predictor to the approximator. 

To cope with this issue, Xu et al.~\cite{Xu2017Iterative} propose to train the approximator and predictor in multiple iterations.
The training process in the first iteration is the same as the onepass training. In the next iteration, they train the approximator using a subset of the input data; the chosen input data was safe-to-approximate in the last iteration. Consequently, the retrained approximator is validated again and generates updated labels for the whole training set, which are used to train the predictor again. Above process repeats iteratively. This training method, when compared to the onepass training, causes more precise approximated results. In fact, the predictor guides the training process of the approximator by selecting the training data. Nevertheless, the interference of the predictor also narrows down the generalization capability of the approximator, who is thereby impotent to the diversified dataset in the field. Their experimental results show data discrimination: two clusters appear in the input data space. In one cluster, the input data leads to much lower approximation error; while the one in the other cluster causes much higher approximation error. As a result, the iterative training method is not designated to improve the invocation---the speedup as well---of the approximate accelerator. 

In previous works, the approximator and predictor are trained separately to minimize their loss functions. The difficulty of finding a joint loss function impedes us to make a good trade-off between the quality and the energy-efficiency. Besides, we have to pay a significant effort and spend much time to search numerous combinations of two sets of hyper-parameters, such as batch size, training rate, and epoch numbers. It is well known in the Machine Learning field that end-to-end training can decrease the supervision needed and balance the training of both NNs. A predictor and an approximator, associated with different tasks, form a composite structure---this is a typical multitask learning scenario. It has been proved that improved generalization error bounds can be achieved because of the shared parameters~\cite{caruana1993multitask,baxter1995learning}. All above motivates us to design a holistic end-to-end trainable neural network for approximate computing with quality guarantee.
\section{Proposed AXNet structure and its training}\label{sect:approach}
Multitask learning can improve generalization by using the domain information contained in the training signals of related tasks as an inductive bias. It does this by learning tasks in parallel while using a shared representation; what is learned for each task can help other tasks be learned better~\cite{caruana1998multitask}. Inspired by this, we train the approximator and predictor in parallel, rather than successively and separately, using a shared representation. To find a shared representation, we first try weight sharing mechanism---a common approach---but fail. Then, we success by fusing the neurons between the approximator and predictor. 
\subsection{Weight sharing mechanism: A false start}\vspace{-2pt}

We first try a commonly used format of shared representation. We use Multi-layer Perceptron~(MLP) as the neural networks in this paper for clarity. We thereby merge the first hidden layers of the approximator and predictor. The rest of the two NNs remain separately, as shown in Figure \ref{fig:weight_sharing}. The resulting neural network contains a prediction subnet and an approximation subnet, inherit the predictor and approximator, respectively.

The training procedure is composed of forward-propagation~(FP) and backward-propagation~(BP). In the FP stage, we apply the input data $\bm{x}$ to the approximation subnet and derive the approximated value $H(\bm{x})$. The approximation error depends on the difference between $H(\bm{x})$ and the exact output $\bm{y}$, as well as the error metric function $Err$, e.g., square-error, etc. We derive the label $\bm{p}_{label}$ by comparing $Err$ with the error bound: 
\begin{equation}
\bm{p}_{label} = bool\lbrace Err(H(\bm{x})-\bm{y}) < error\ bound\rbrace
\label{eqt:plabel}
\end{equation}
$\bm{p}_{label}$ tells whether the input data is safe-to-approximate, and is further used in the cost function~(\ref{eqt:costfunction}) to train the prediction subnet. 
\begin{equation}J =  L_{a}(\bm{y}, H(\bm{x})) + L_{p}(\bm{p}_{label},P(\bm{x}))
\label{eqt:costfunction}
\end{equation}
wherein $P(x)$ is the output of prediction subnet indicating the classification result, $L_{a}$ denotes the loss function of the approximation subnet and $L_{p}$ refers to the loss function of the prediction subnet, i.e., the cross entropy. 
In the BP stage--using Stochastic Gradient Descendent algorithm--~(two sets of) gradients originated from two different loss functions,  $L_{a}$ and $L_{p}$, pass through all hidden layers of two subnets separately until reaching the layer with shared neurons. The sums of the two gradients are used to update the shared neurons.

\begin{figure*}[htbp]\centering
\begin{minipage}{0.28\linewidth}
\centering
    \includegraphics[width=1\linewidth]{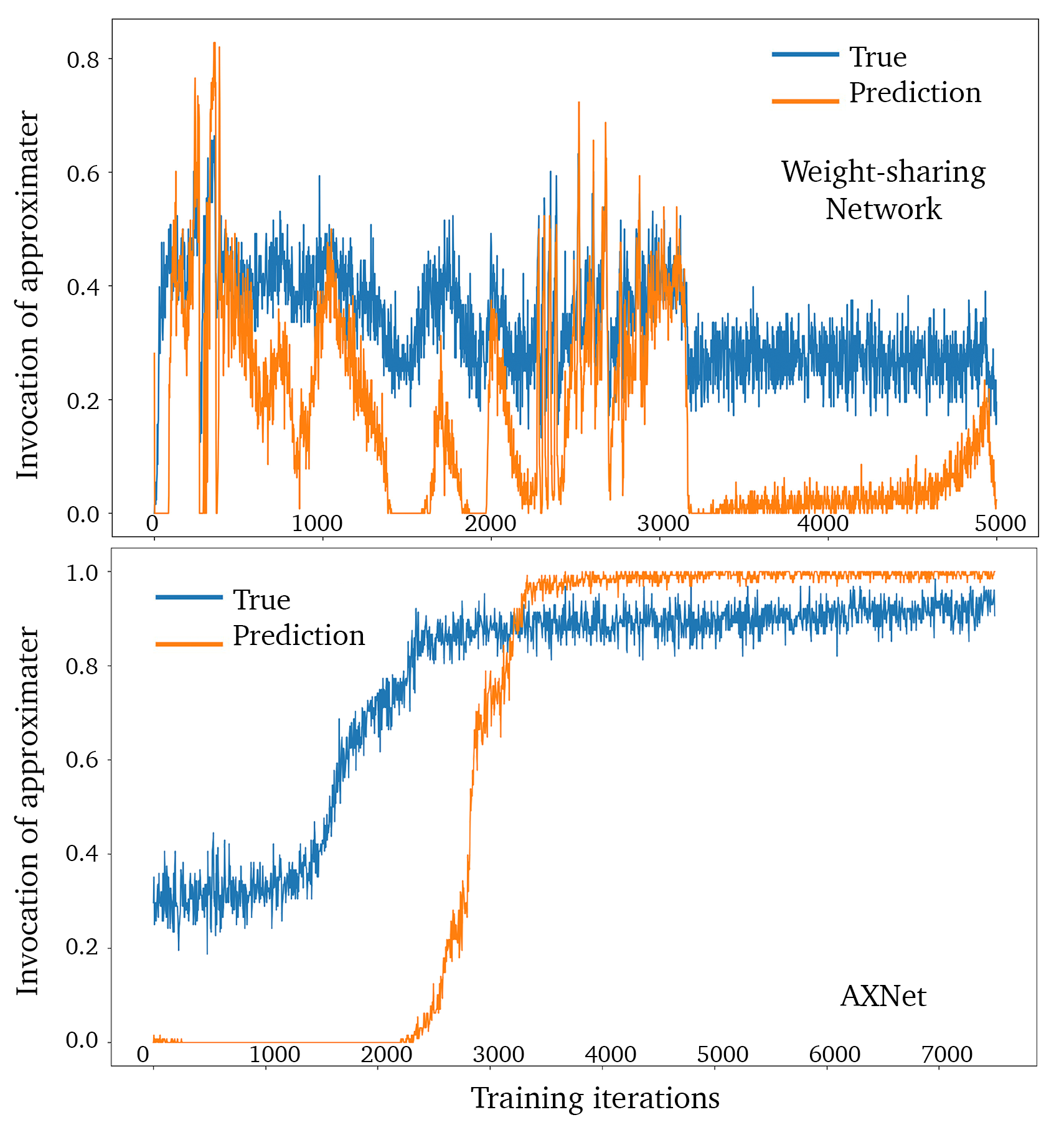}
    \caption{Change of invocation in training weight-sharing network (Up) and AXNet (Down).}
    \label{fig:oscillation_of_two_networks}
    \end{minipage}\hfill
    \begin{minipage}{0.66\linewidth}
    \centering
    \includegraphics[width=\linewidth]{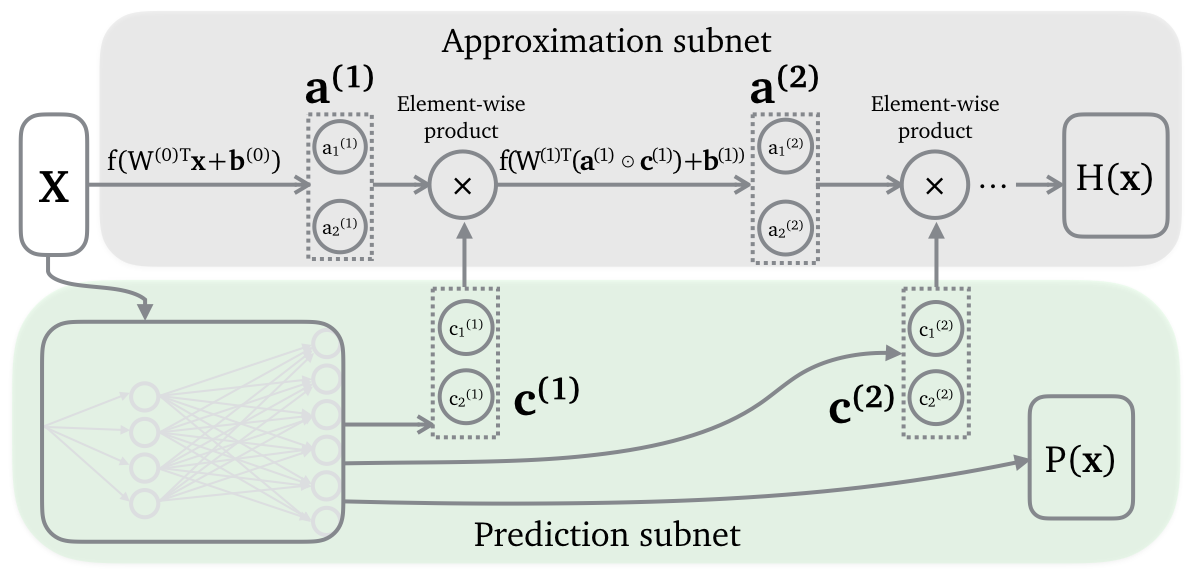}
    \caption{Structure of AXNet. Note that $\bm{c}^{(1)}, \bm{c}^(2),..., P(\bm{x})$ are split from the output layer of prediction subnet. Do not consider these vectors as different layers.}
    \label{fig:AXNet}
        \end{minipage}
\end{figure*}

Such neural network is end-to-end trainable but has a highly unstable training process, which always converges to a low invocation. Figure~\ref{fig:oscillation_of_two_networks}(a) provides a preliminary experiment by training such a weight-sharing neural network. 
We find that the gradient of the prediction subnet~(Cross-Entropy) is, in most cases, an order of magnitude higher than that of the approximation subnet~(MSE), but their difference varies with time. Consequently, the gradients of the prediction subnet dominate the update of all the shared weights. 
At the beginning of the training procedure, the invocation of the approximation subnet significantly causes the turbulence of prediction results $\bm{p}_{label}$, resulting in a drastic change of $L_{p}$. We then observe a significant fluctuation of the shared weights that aggravate swing of the invocation of the approximation subnet. Such interference between two subnets always leads to two controversial gradients before updating the shared weights, which in turn incurs the oscillation in the training procuedure.
We cannot diminish this phenomena by scale the gradients due to the ignorance of the exact order of magnitude of these two gradients.

\subsection{Structure and training of AXNet}\vspace{-2pt}
To avoid above coupling effect between the two subnets, in this section, we describe our proposal \textbf{AXNet}. 
The structure of AXNet is shown in Figure~\ref{fig:AXNet}.

Consider an approximation subnet which has an input vector $\bm{x}$ with size $N$, and $l$ hidden layers. Hidden layer $i$ has $L_i$ neurons and outputs a vector of activation values, $\bm{a}^{(i)}$. $H(\bm{x})$ denotes the approximated values.
The prediction subnet has an output layer $\bm{c}$.
We split $\bm{c}$ into $l+1$ vectors:
$$[\bm{c}^{(1)}, \bm{c}^{(2)}, ..., \bm{c}^{(l)}, P(\bm{x})]$$
First $l$ vectors, called \textbf{control vector}, ``control'' the approximation subnet. Last vector $P(\bm{x})$ is the prediction result and has one value if we apply simoid function in the preceding neurons, or two values when applying softmax activation. 
$\bm{p}_{label}$ and $J$ are defined identically as in equation~(\ref{eqt:plabel}) and~(\ref{eqt:costfunction}).
Note that the output layer of the prediction subnet requires $N_{o}$ neurons, depending on the choice of activation function for $P(\bm{x})$, i.e., softmax or sigmoid.
\begin{equation}
N_{o} = \left\{
\begin{array}{lr}
\sum_{i=1}^{l} L_i + 2 &\text{if softmax}\\
\sum_{i=1}^{l} L_i + 1  &\text{if sigmoid}
\end{array}
\right.
\end{equation}

The essence of AXNet is carrying out the Hadamard product (denote as "$\odot$") between the activation vector $\bm{a}^{(i)}$ and the corresponding control vector $\bm{c}^{(i)}$. The resulting vector $\bm{a}^{(i)}  \odot \bm{c}^{(i)}$ is passed to the successive layers acting as input vector in the approximation subnet. Namely: 
\begin{equation}
\bm{a}^{(i+1)} = f(W^{(i)T}\bm{a}^{(i)}\odot \bm{c}^{(i)} + \bm{b}^{(i)})
\label{eqt:product}
\end{equation}
wherein $f$ denotes the activation function of hidden layer $i$. Consequently, all hidden layers of approximation subnet interlink with the output layer of prediction subnet. 
 \RestyleAlgo{boxruled}
 \SetKwProg{Fn}{Function}{}{}
 \IncMargin{1em}
 \begin{algorithm}[!htp]
     \caption{Training Procedure of AXNet.}
          \label{alg:train}
     \SetKwInOut{Input}{Input}
     \SetKwInOut{Output}{Output}
     \Input{$\bm{X}$: The input features of training data \newline 
     $\bm{Y}$: The fitting target of training data \newline
     $l$: Num. of hidden layers in approximation subnet \newline
     $MaxIterations$: maximum iterations
     }
     \While {$iterations < MaxIterations$}{
              $\bm{x},\bm{y} \gets \text{a batch of training samples}$\;
           \emph{\# Forward propagation.\\}
         Get $[\bm{c}^{(1)}, ..., \bm{c}^{(l)}, \bm{p}] $ by feeding prediction subnet $\bm{x}$.\\
         $\bm{h}^{(0)} = \bm{x}$\;
         \For{i=1,...,l}
         { \emph{\# Inference of approximation subnet.}\\
        $\bm{a}^{(i)} = \bm{b}^{(i)} + W^{(i)T}\cdot \bm{h}^{(i-1)}$\;
        $\bm{h}^{(i)} = f(\bm{a}^{(i)})\odot \bm{c}^{(i)}$\;
         }
         Calculate $H(x), \bm{p}_{label}, L_a, L_p$.\\
         \emph{\# Backward propagation.}\\
         $\bm{g}=\Delta_{H(x)} L_a$;\\
         \For{k=l,l-1,...,1}
         {\emph{\# Update approximation subnet.}
         $\bm{g}= \Delta_{\bm{a}^{(k)}}L_a = \bm{g}\odot f'(\bm{a}^{(k)})$\;
         $\Delta_{\bm{b}^{(k)}}L_a=g$\;
          $\Delta_{\bm{W}^{(k)}}L_a=\bm{g}\bm{h}^{(k-1)T}$\;
          Update $\bm{b}^{(k)}$ and $\bm{W}^{(k)}$\;
          $\bm{g}=\Delta_{\bm{h}^{(k-1)}}L_a = \bm{W}^{(k)T} \bm{g} \odot \bm{c}^{(k)}$\;
         }
         Calculate all $\Delta_{\theta} (L_a + L_p)$ and update all $\theta$ for parameters $\theta$ in prediction subnet.\\
     }
 \end{algorithm}
 \DecMargin{1em}

The entire network can still be trained in an end-to-end manner in back-propagation. 
The algorithm is shown in Algorithm~\ref{alg:train} (refer to algorithms 6.3, 6.4 in \cite{goodfellow2016deep}). A batch of training samples pass through prediction subnets to collect control vector (line 4). Then FP of approximation subnet (line 5-11) derives $\bm{p}_{label}$ for training prediction subnet. In line 9, we apply Hadamard product to the activation value. In the BP stage (line 12-21), the gradients of $L_a$ pass through approximation subnet and the gradients of both $L_a$ and $L_p$ are used for updating the prediction subnet.

AXNet shows excellent training stability and convergence rate, as shown in Figure~\ref{fig:oscillation_of_two_networks}(b). Interestingly, the convergence of the prediction subnet falls behind that of the approximation subnet. We denote this phenomenon as ``saturation effect'', and attribute the successful training of AXNet to this saturation effect. The cause of saturation effect is the skewness of $\bm{p}_{label}$ provided to train prediction subnet when the true invocation of approximation subnet is near  $0\%$ or $100\%$. According to previous study~\cite{lawrence1998neural}, if the number of training examples that correspond to each class---safe-to-approximate or not---varies significantly between the classes, then it may be harder for the network to learn the rarer classes in some cases. Thus in the beginning, the prediction subnet fails to catch up with the immature approximation subnet. Different from the weight-sharing method, the failed training of prediction subnet does not affect the training of the approximation subnet because the approximation subnet is relatively independent of the prediction subnet. Unfortunately, this property also damages the performance of AXNet when the approximation subnet is invoked almost $100\%$. Under this circumstance, common techniques to tackle imbalance data can be used \cite{longadge2013class}.

Note that, previous works~\cite{Xu2017Iterative,Mahajan2016Towards} train the predictor sufficiently \emph{after} the approximator. All these works, including this work, provide the evidence to advocate the delay~(less effort) of training the predictor in the beginning of the training process, when the approximator is too weak to provide a high-quality approximate output. Otherwise, the skewed samples~(most of them are unsafe-to-approximate) will destroy the training of the predictor. The resulting predictor makes inaccurate, if not absurd, predictions on data, which in turn misleads the training of the approximator. The same phenomena can be observed in training a Generative Adversible neural network~(GAN)~\cite{2016GAN}. A common trick is to train the generator less frequently than train the discriminator.

\subsection{Analysis and Interpretation}
\label{chap3c}
Besides the training stability, we mathematically prove other superior properties of AXNet:

First, AXNet improves the capacity of fitting the target function by introducing extra non-linearity using the Hadamard product operations. 
Without loss of generality, suppose both the prediction subnet and the approximation subnet are MLPs with linear activation function. By rewriting the input vector that passes to a hidden layer of approximation subnet in a concrete mathematical form, we derive the Hadmard product $\bm{a}^{(1)} \odot \bm{c}$ (which is sent to next layer in approximation subnet) in dimension $i$:
\begin{equation}
    \begin{split}
    (\bm{a}^{(1)} \odot \bm{c})_i = & \bm{a}^{(1)}_i \cdot \bm{c}_i\\
        = & (\sum_{j=1}^{N}W^{(a1)}_{(i,j)}\bm{x}_j + b^{(a1)}_{i})
    \cdot
    (\sum_{k=1}^{N}W^{(c)}_{(i,k)}\bm{x}_k + b^{(c)}_{i})\\
        = & \sum_{k=1}^{N} \sum_{j=1}^{N}W^{(a1)}_{(i,j)}W^{(c)}_{(i,k)}\bm{x}_j \bm{x}_k\\
        & + \sum_{j=1}^{N}(b^{(c)}_{i}W^{(a1)}_{(i,j)}+b^{(a1)}_{i}W^{(c)}_{(i,j)})\bm{x}_{j}\\
        & + b_i^{(c)} b_i^{(a1)}\\
    \end{split}
\end{equation}
wherein $\bm{x}$ is $N$-dimensional input vector, $W$ refers to the weight matrix, $\bm{b}$ is the bias vector, $x_i, b_i$ refers to the $i_{th}$ value of the vector $\bm{x},\ \bm{b}$, respectively. $W^{(layer)}_{(i,j)}$ denotes the element of the weight matrix in $layer$. 
This equation tells that the combination of all input features, namely quadratic terms, are passed to the rest of the approximation subnet. The successive hidden layers have even higher order terms. The Hadamard product thereby introduces the higher order terms, extra non-linearity and more complex representation of input features. High order terms of input features have been widely used in previous machine learning practices as feature engineering technique~\cite{zabokrtsky2015feature}, but by hand-craft selecting instead of automatically generating like this work.

\begin{figure}[htbp]
\begin{centering}
\subfigure[Output surface of bessel function.]{
\centering
    \includegraphics[width=0.45\linewidth]{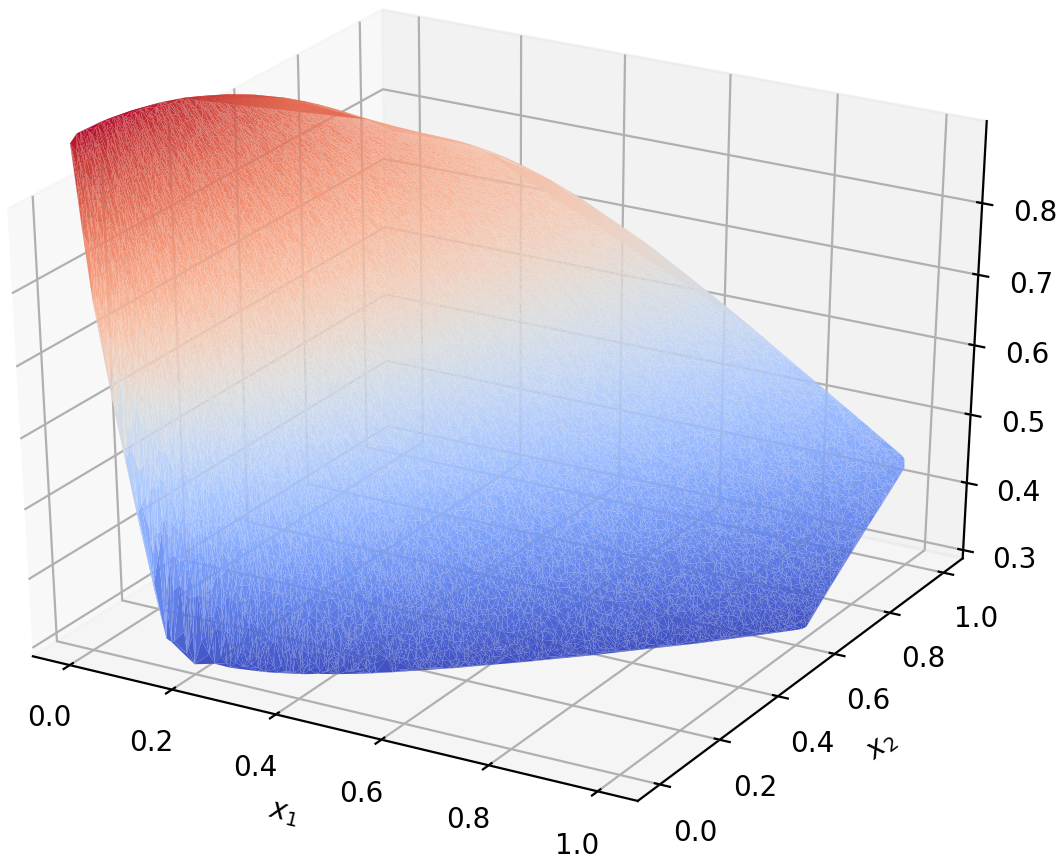}
    \label{fig:3doriginal}}
\subfigure[Output surface of AXNet.]{\centering
    \includegraphics[width=0.45\linewidth]{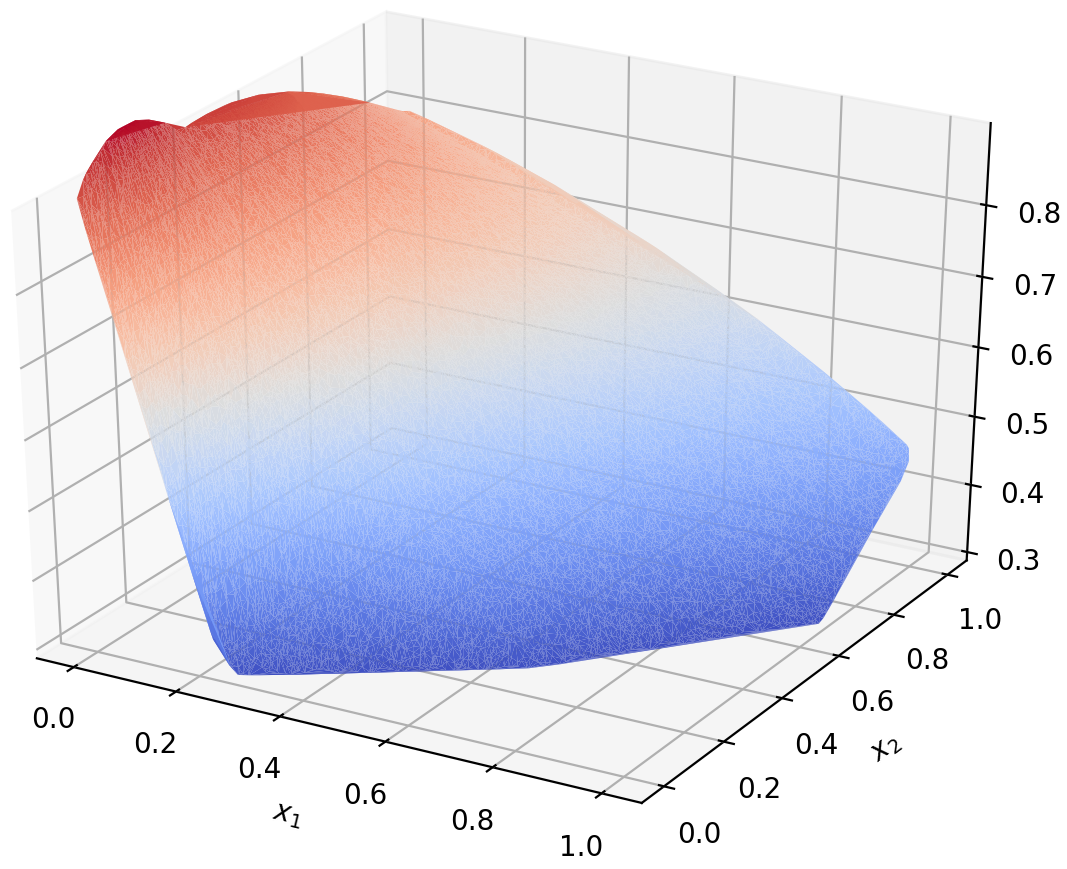}
    \label{fig:3daxnet}}
\end{centering}
~\\
\begin{centering}
\subfigure[Output surface of single approximator.]{\centering
    \includegraphics[width=0.45\linewidth]{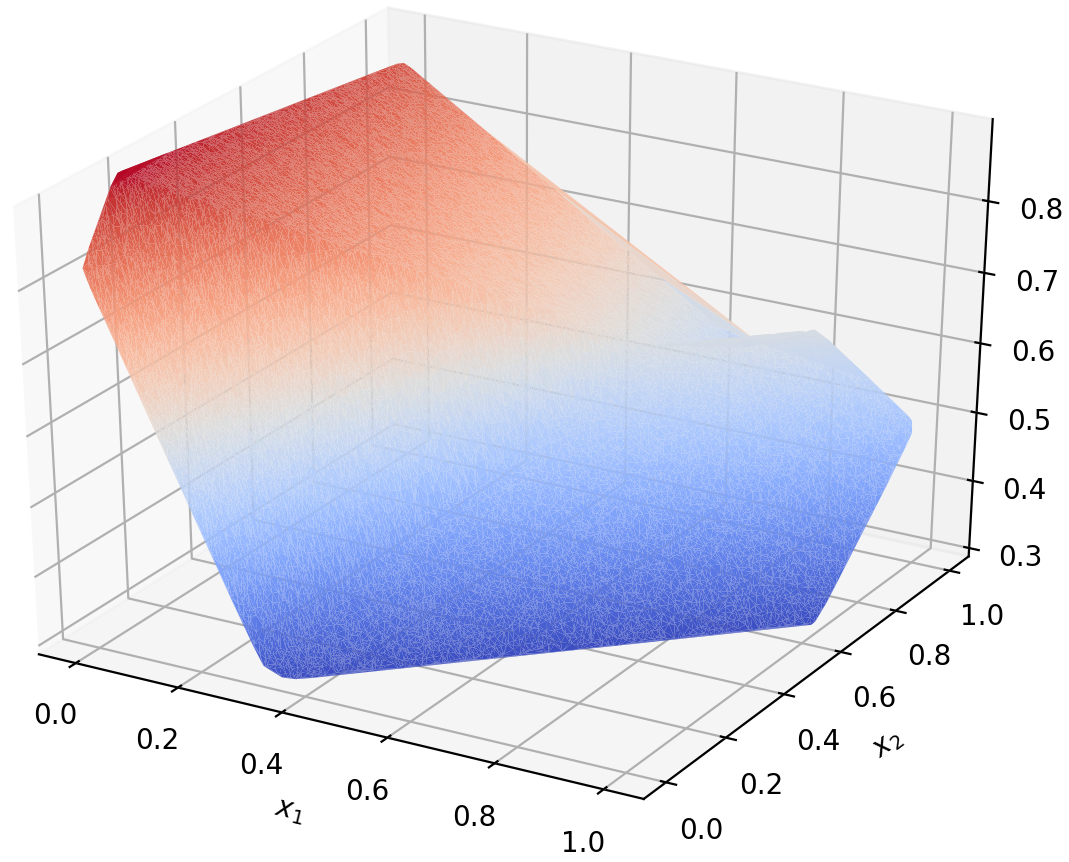}
    \label{fig:3dsingle}}
\subfigure[Activation value before and after Hadamard product.]{\centering
    \includegraphics[width=0.45\linewidth]{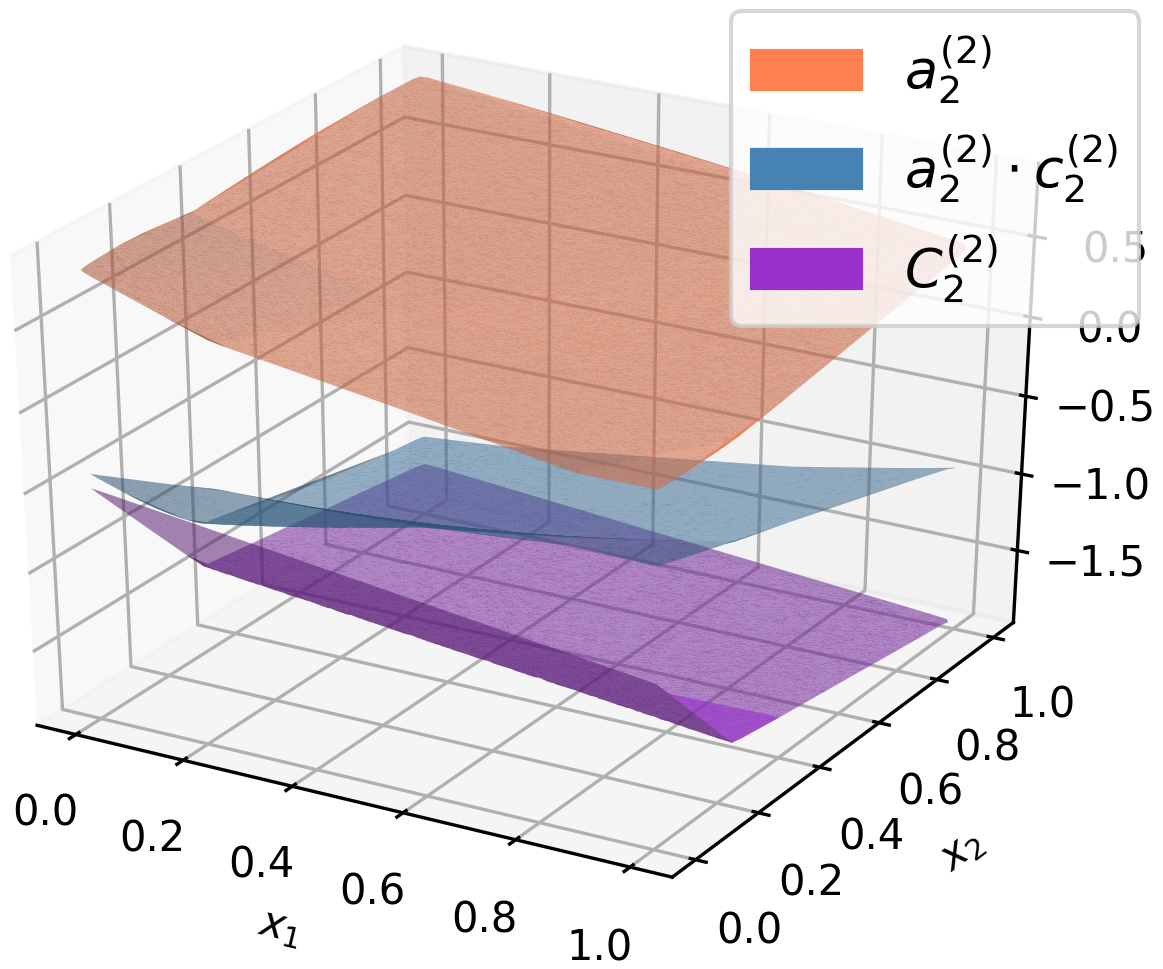}
    \label{fig:3dcontrol}}
\end{centering}
\caption{A case study to illustrate the adaptive adjustment imposed to the approximation subnet by control vector. When $(x_1, x_2)\to (0, 1)$, we can see the control vector suppresses the activation value.}    
\label{fig:beforeafter}
\end{figure}
Second, thanks to the control vectors, AXNet can adjust the activation values of the hidden layers of the approximation subnet. Control vectors filter the activation value of the hidden layers in the approximation subnet through Hadamard product. Fig.~\ref{fig:beforeafter} demonstrates a case study of this effect. Bessel function is suitable for visualization as it has two-dimensional inputs, drew in $x_1, x_2$ axis, and one-dimensional output, drew in vertical axis. These figures show the existence of the prediction subnet improves the fitting capacity of the approximation subnet. When the input~($x_1, x_2$) approaches to the corner~($0, 1$), the control value $\bm{c}^{(2)}_{2}$ in vector $\bm{c}^{(2)}$ suppresses the activation value $\bm{a}^{(2)}_{2}$ (See Fig \ref{fig:3dcontrol}). Same effect happens at other neurons in this layer. This is the reason AXNet (Figure \ref{fig:3daxnet}) produces better result than single approximator (Figure \ref{fig:3dsingle}) near the corner~$(x_1, x_2)=(0,1)$.

Third, this end-to-end network structure inherently balances the two learning tasks and seeks the global optimism due to the joint loss function in equation \ref{eqt:costfunction}. When we train two isolated NNs, each of them inevitably seeks for their respective optimal parameters. However, AXNet enforces one subnet considers the loss of the other during the training procedure. Two subnets thus coordinate to achieve the minimal loss, resulting in the maximal invocation and the minimal approximation error. This coordination is more effective than the one in \cite{Xu2017Iterative} by selecting the training samples.

\subsection{Subnet fusion with single control vector}
A drawback of the current AXNet design is the increased neurons and synaptic weights in the output layer of prediction subnet as well as the extra Hardmard production. We need $\sum_{i=1}^{l} L_i $ extra neurons for control vectors.
If the approximation subnet become larger, the cost of AXNet is larger. 

To  resolve the above issue, we further orchestrate a simpler AXNet by interlinking prediction subnet with a single hidden layer, instead of all the hidden layers, of approximation subnet. In that hence, the prediction subnet only need a single control vector, which dramatically reduce the storage and computation overhead. The experimental result confirms that the simplified AXNet maintains its performance ~(Figure~\ref{fig:scale}). 
\section{Architecture of AXNet Accelerator}\label{sect:architecture}
\begin{figure}[bth]
\subfigure[]{
    \centering
        \includegraphics[width=0.61\linewidth]{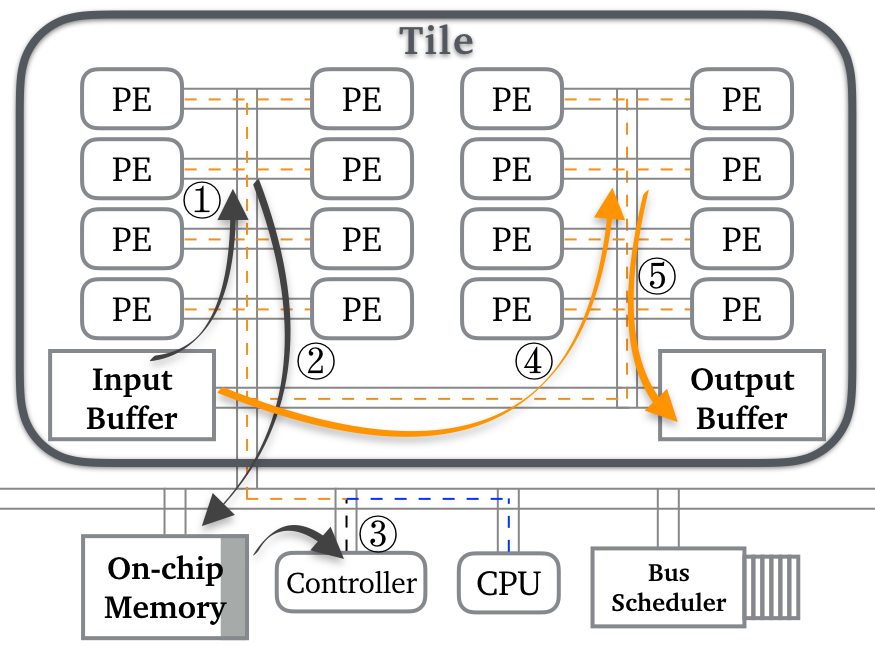}
    \label{fig:architecture}}
\subfigure[]{
    \centering
    \includegraphics[width=0.32\linewidth]{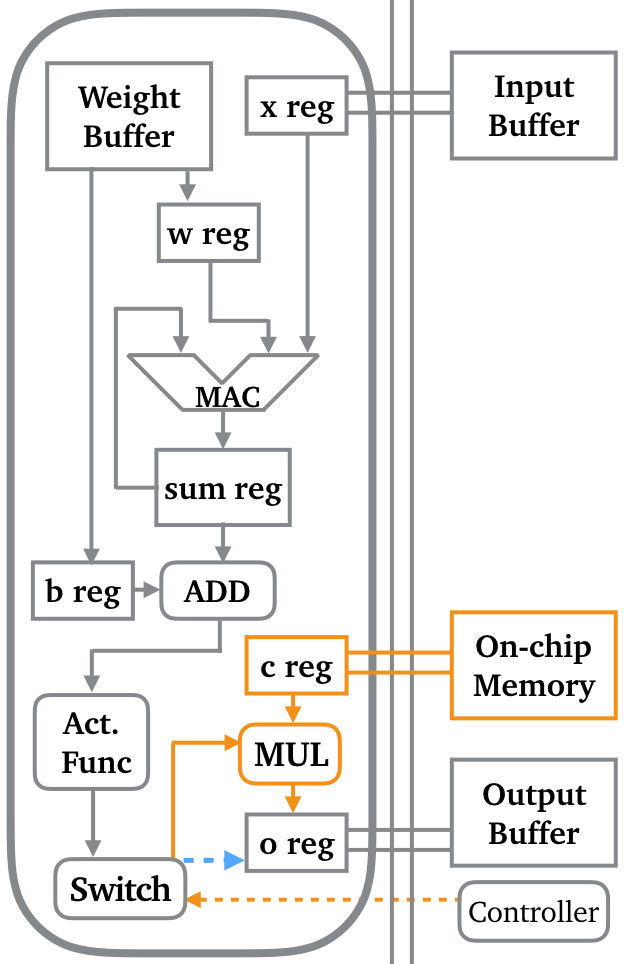}
    \label{fig:inner}
}
\caption{Proposed NPU architecture. (a)The structure of NPU and the data flow of AXNet. (b)The Structure of PE.}
\end{figure}

Due to the space limit, this section describes an simple NPU architecture, imitating the NPU architecture in~\cite{hadi2012neural}, which fits AXNet. As shown in Figure \ref{fig:architecture}, the NPU contains many processing engines~(PEs), grouped as Tiles, a controller, an on-chip memory, and a bus scheduler. A tile~(encircled by the rounded rectangle) is composed of a set of identical PEs, an input buffer and output buffer, all of which are connected by an internal bus~(we omit the internal bus arbiter for clarity). We adopt neuron-level parallelism. Thus, each PE computes the output of a single neuron~(as equation~\ref{eqt:product}) in the prediction subnet or approximation subnet. The input /output buffer temporally stores the input/the output vector, and interfaces with the on-chip memory. The on-chip memory can interface with the DRAM, input/output buffers in the tile and CPU through the bus. It can store the weight matrix, the input samples and output results of AXNet, and the intermediate results transferring between two adjacent layers in the neural network. The controller is responsible for sending invoke signal through control bus~(dotted lines) to PEs or CPU according to the prediction result~(i.e., $P(\bm{x})$). As data transfers concurrently between tiles, CPU and on-chip memory, a bus scheduler is necessary to avoid bus conflict.

Figure \ref{fig:architecture} shows the data flow of executing AXNet. When the input sample comes into the on-chip memory, the NPU schedules the computation for prediction subnet~(the first three stages) and subsequently for approximation subnet~(the last two stages). In stage \textcircled{\footnotesize{1}}, the input data $\bm{x}$ is fetched from the on-chip memory to the Input Buffer through the data bus. The weight buffer in each PE fetches the weight vectors from the on-chip memory. When receiving both input and weight vector, each PE conducts forward propagation of prediction subnet and generates prediction result $P(\bm{x})$ and control vectors $\bm{c}^{(i)}$. In stage \textcircled{\footnotesize{2}}, the control vector and the prediction result $P(\bm{x})$ of the prediction subnet are sent to the on-chip memory. Specifically, $P(\bm{x})$ is placed in a specified address~(the grey region inside of the on-chip memory). In stage \textcircled{\footnotesize{3}}, the controller gets $P(\bm{x})$ from the on-chip memory. According to the value of $P(\bm{x})$, the controller invokes either the CPU or the approximation subnet through control bus~(dotted lines). If approximation subnet is invoked, in stage \textcircled{\footnotesize{4}}, each PE fetches input data and control vectors from buffers and conduct forward propagation of approximation subnet. In stage \textcircled{\footnotesize{5}}, the approximation result is sent to the on-chip memory through the data bus.

Note that, the proposed NPU can statically allocate the computing/storage resource for the whole AXNet if the derived AXNet for an application is small enough. In this case, the weight vector can stay in the weight buffer of each PE all the time. Otherwise, the NPU can dynamically schedule the computation of AXNet layer by layer. In that hence, the input/output buffer of each PE and the on-chip memory will temporarily accommodate the intermediate results between adjacent layers.

We modify a general PE to compute the Hadamard product induced by the fusion of two neural networks. Figure \ref{fig:inner} shows the internal structure of such PE. When the PE loads the input vector into \textbf{x reg} from \textbf{Input Buffer} and loads weight data into \textbf{w reg} and \textbf{b reg} from the \textbf{Weight Buffer}, the \textbf{Multiply Add Unit} calculates the dot product of the input vector and the weight vector. The resultant product is stored in the \textbf{temporary reg}. After adding the bias, the result is sent to the \textbf{Activate Unit}, which implements the activate function~(i.e., relu). Though NPU performs different computations in prediction subnet and approximation subnet, the PE has the same structure leveraging a \textbf{switch unit} after \textbf{Activate Unit}. At first, the \textbf{switch unit} enables the blue dotted path which directly pushes the activation result into the \textbf{output reg}. When receiving an invoke signal in stage \textcircled{\footnotesize{3}}, indicating the computation of approximation subnet,  the \textbf{switch unit} activates the approximation subnet units~(in solid orange lines). Inside of the PE, the Hadamard product reduces to a standard multiplication operation between one activation value, say $\bm{a}^{(i)_j}$, and the corresponding element in control vector $\bm{c}^{(i)}_{j}$. Therefore, a \textbf{Multiplier} can carry out the above computation. The output of  $\bm{c}^{(i)}_j\times \bm{a}^{(i)}_j$ is stored in \textbf{o reg} and waits for the transferring to \textbf{Output Buffer}.
\section{Experiments}\label{sect:experiment}
\begin{figure*}[b]
\begin{minipage}[b]{0.6\linewidth}
\centering
\includegraphics[width=\linewidth]{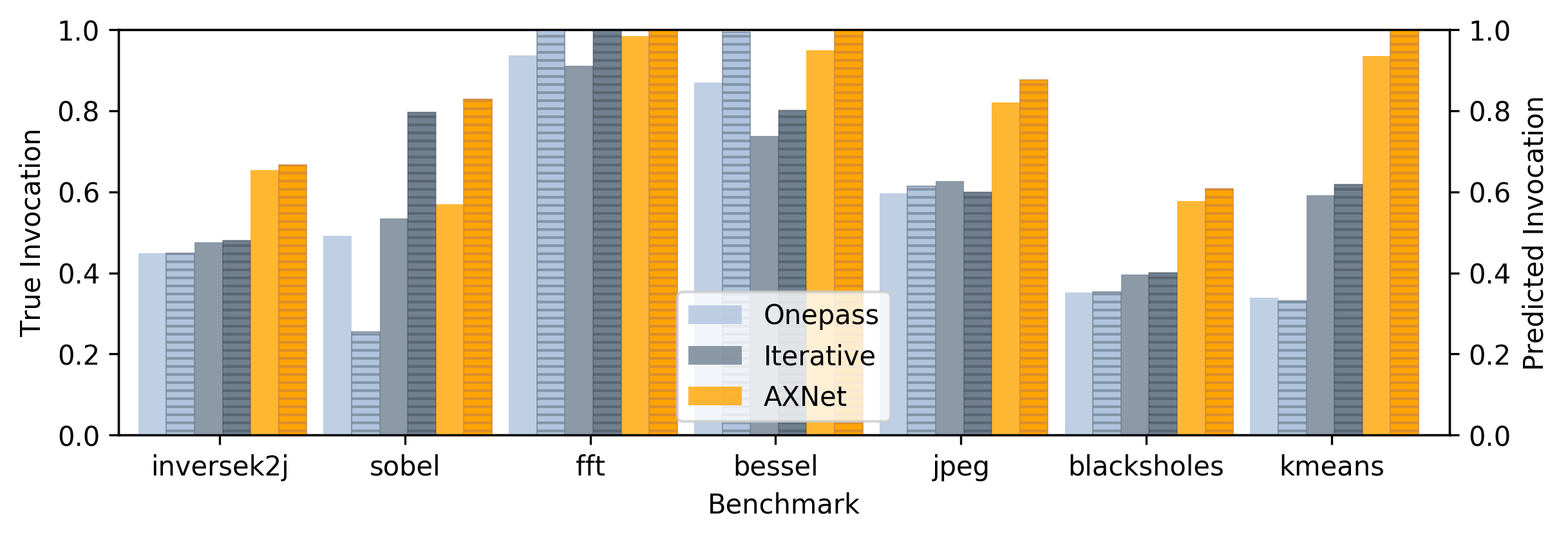}
\caption{Comparisons on the true invocation (solid color bars) and predicted invocation (bars with grey lines).}
\label{fig:invo_all}
\end{minipage}
\begin{minipage}[b]{0.36\linewidth}
    \centering
        \includegraphics[width=0.9\linewidth]{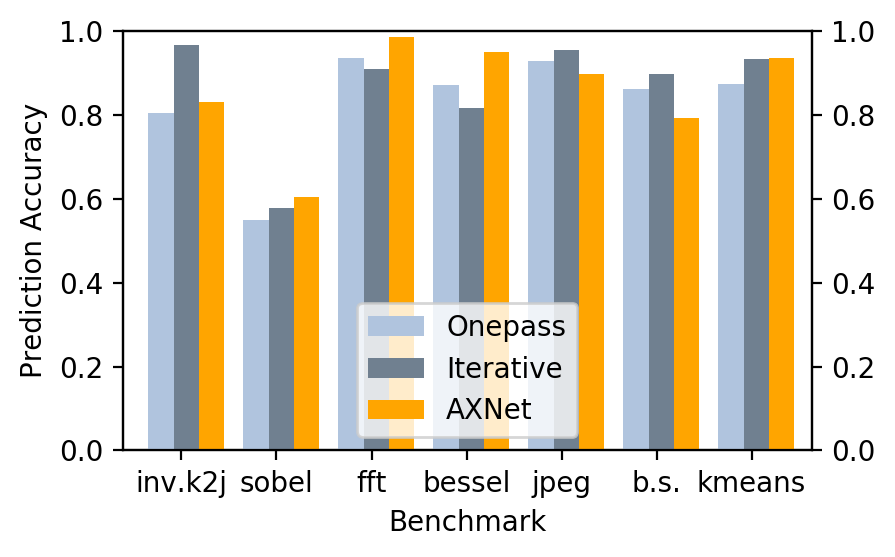}
        \caption{Comparisons on the prediction accuracy}
    \label{fig:accuracy}
    \end{minipage}
\end{figure*}

\subsection{Experimental Setup}\vspace{-2pt}
We compare the proposed AXNet to two typical previous methods, i.e., ``onepass'' \cite{Mahajan2015neuralprediction} training and ``iterative'' \cite{Xu2017Iterative} training, using identical optimizer~(i.e., Adam), the same error metrics, error bound, number of hidden layers, activation function~(i.e., ReLu) and loss functions~(MSE for approximation and cross entropy for prediction). We choose target functions from a widely used dataset for approximate computing, AxBench, including FFT, Bessel, Blackscholes, jpeg, inversk2j, kmeans and Sobel~\cite{2016axbench}. AxBench provides tremendous amount of the training data and testing data.
Note that we choose these benchmarks because first, they are typical applications covering predominant domains in approximate computing, and second, these choices follow the path of previous works \cite{Mahajan2016Towards,Xu2017Iterative,He2016ACR}.

AXNet has a similar structure and approximately equal parameter count as the neural networks in previous methods. However, AXNet introduces more parameters in the output layer of the prediction subnet.
For a fair comparison, we compare i)the invocation by shrinking the structure of AXNet to match the parameter count~($\le 3\%$ difference) of the neural networks in previous methods, and ii)the parameter count by permuting the AXNet to match others' invocation~($<5\%$ difference). Table~\ref{tab:bench} shows all experimental setup.
\begin{table}[tb]
    \centering
    \resizebox{\linewidth}{!}{
    \begin{tabular}{|c|c|c|c|c|c|c|}
      \hline
      \textbf{Benchmark} &
      \textbf{Domain}&
      \tabincell{l}{\textbf{Error bound \&}\\\textbf{ Error metrics}}&
      \textbf{Method}&
      \textbf{A Topology} &
      \textbf{P Topology} &
      \tabincell{c}{\textbf{Para.}\\\textbf{count}}\\\hline
      
     \multirow{3}{*}{inversek2j}  &
      \multirow{3}{*}{Robotics}  &
     \multirow{3}{*}{\tabincell{c}{0.01\\relative error}}&
     \multirow{2}{*}{AXNet}    &
     2-6-2 &
     2-4-8  &
     84\\\cline{5-7}
     
     &&&&
     2-5-2&
     2-3-7&
     64\\\cline{4-7}

     &&
     & previous &
     2-8-2 &
     2-8-2  &
     84\\\hline
        
    \multirow{3}{*}{sobel}&
     \multirow{3}{*}{\tabincell{c}{Image\\Processing}}  &
    \multirow{3}{*}{\tabincell{c}{0.01\\image diff} }&
    \multirow{2}{*}{AXNet}&
    9-8-1&
    9-3-10&
    159\\\cline{5-7}

	&&
	&
	&9-7-1&
	9-3-9&
	144\\\cline{4-7}
	
    &&&previous&
    9-8-1&
    9-8-2&
    187\\\hline
    
    \multirow{2}{*}{FFT}&
     \multirow{2}{*}{\tabincell{c}{Signal\\Processing}}&
    0.05&
    AXNet&
    1-4-3-2&
    1-3-9&
    41\\\cline{4-7}
    
    &&
    absolute error&previous&
    1-4-4-2&
    1-4-2&
    56\\\hline
    
      \multirow{3}{*}{bessel}&
      \multirow{3}{*}{\tabincell{c}{Scientific\\computing}}&
    \multirow{3}{*}{\tabincell{c}{0.05\\absolute error} }&
      \multirow{2}{*}{AXNet}&
      2-2-2-1&
      2-4-6&
      57\\\cline{5-7}
      
    &&
    &&2-2-2-1&
    2-2-6&
    39\\\cline{4-7}
    
    &&&previous&
    2-4-4-1&
    2-4-2&
    59\\\hline
    
      \multirow{3}{*}{jpeg}&
      \multirow{3}{*}{\tabincell{c}{Image\\processing}}&
      \multirow{3}{*}{\tabincell{c}{0.001\\image diff}} &
      \multirow{2}{*}{AXNet}&
      64-16-64&
      64-12-18&3129\\\cline{5-7}
      
      &&&&64-6-64&
      64-6-8&
      1284\\\cline{4-7}
      
      &&&
      previous&
      64-16-64&
      64-16-2&
      3216\\\hline
      
\multirow{3}{*}{blackscholes}&
\multirow{3}{*}{\tabincell{c}{Financial\\analysis}}&
  \multirow{3}{*}{\tabincell{c}{0.001\\relative error}} &
  \multirow{2}{*}{AXNet}&
  6-6-1&
  6-4-7&
  112\\\cline{5-7}
  
  &&&&6-5-1&
  6-3-7&
  90\\\cline{4-7}
  
      &&&
   previous&   6-8-1&
      6-8-2&
      138\\\hline
      
\multirow{3}{*}{kmeans}&
\multirow{3}{*}{\tabincell{c}{Machine\\learning}}&
  \multirow{3}{*}{\tabincell{c}{0.01\\image diff}} &
  \multirow{2}{*}{AXNet}&
  6-4-4-1&
  6-4-10&
  131\\\cline{5-7}
  
  &&&&6-3-2-1&
  6-3-7&
  81 \\\cline{4-7}
  
      &&&
   previous&   6-4-4-1&
      6-8-2&
      127\\\hline

    \end{tabular}
    }
    
    ~\\
    \caption{Experimental setup of all benchmarks.}
    \label{tab:bench}\vspace{-10pt}
\end{table}

We used four evaluation metrics defined as follows:
\begin{itemize}
    \item \textbf{True invocation}: the proportion of safe-to-approximate samples among all the testing data. 
    \item \textbf{Predicted invocation}: the proportion of samples that the prediction subnet believes to be safe-to-approximate among all testing data.  
    \item \textbf{Prediction accuracy}: the proportion of samples that are safely approximated meanwhile predicted as safe-to-approximate.
    \item \textbf{Approximation error}: the mean error of approximation results for those predicted safe-to-approximate samples, also called ``overall error''. Concretely, it's $E[Err(H(\bm{x})-\bm{y})]$ for all $\bm{x}$ labeled safe-to-approximate by prediction subnet.
\end{itemize}
True invocation evaluates the ability of approximation subnet and the approximators in previous works. Predicted invocation and prediction accuracy measure the performance of prediction subnet and the predictor/classifier in previous works. Approximation error assess the overall performance of the approximate computing framework.

To evaluate the energy-efficiency, we first derive the speedup and energy reduction of AXNet and then obtain the improvement of energy-efficiency by $\frac{Speedup}{Energy\ reduction}$. 
Due to the space limit, we theoratically estimate the performance of AXNet accroading to the performance of NPU in \cite{hadi2012neural}, which is valid due to the proposed AXNet is merely the same as the original work in terms of NPU design. The extra overhead of controller, the light modification of PE can be ignored in a NPU design.

\subsection{Result and analysis}\vspace{-2pt}

Figure~\ref{fig:invo_all} shows the true and predicted invocation across different benchmarks. AXNet achieves greater true invocation than iterative and onepass methods in all benchmarks by 30.8\% and 50.7\% respectively in average. The greatest improvement in kmeans and jpeg benchmark take $1.5\times$ and $1.3\times$ respectively compared to the iterative method. AXNet also outnumbers other methods in terms of predicted invocation. Saturation effect happens in benchmark fft, bessel, and kmeans (discussion is at Chapter \ref{chap3c}), resulting in 100\% predicted invocation.

Figure~\ref{fig:accuracy} shows the prediction accuracy. Although our end-to-end trainable AXNet is not sufficiently trained as iterative training, it shows a similar prediction accuracy. In some cases, like inversek2j, jpeg, and blackscholes, the classifier (the same as ``predictor'' in this paper) with iterative training outnumbers AXNet's prediction accuracy. This is because the iterative training method selects the training data in favor of the classifier. In practice, AXNet still carries out more acceptable approximation than the previous methods due to the higher predicted invocation. 

\begin{figure}[tb]
            \centering    \includegraphics[width=0.8\linewidth]{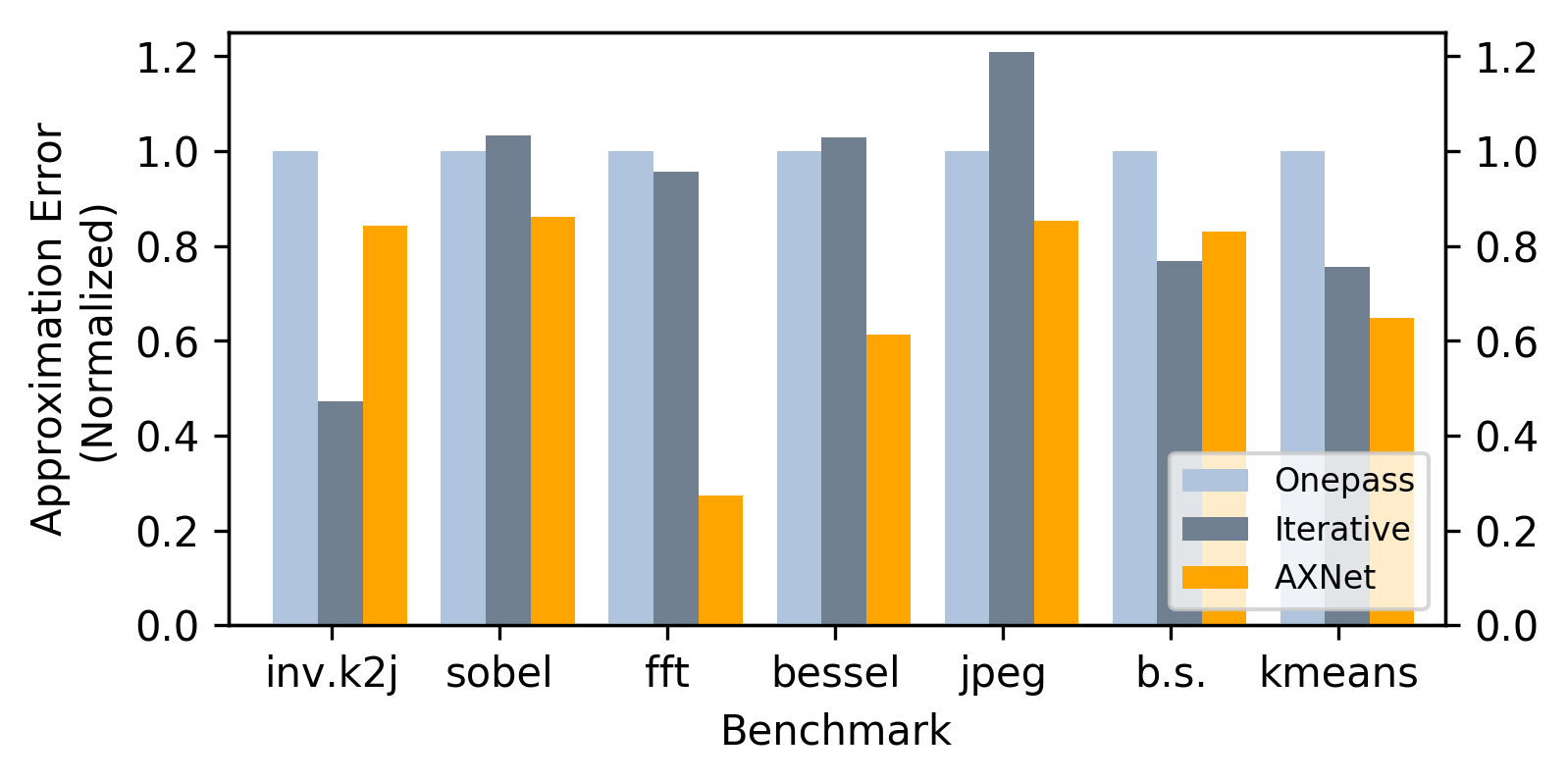}
        \caption{Comparisons on the overall approximation error}
        \label{fig:error}
\end{figure}

Figure~\ref{fig:error} presents the overall approximation error that the user finally observes. In each benchmark, we normalize the overall error to that of the onepass method. In all cases, AXNet has an excellent reduction of error compared to onepass method but in some cases falls behind the iterative method. Note that the overall error is already under the error-bound and has no impact on the quality of approximate computing. 
\begin{figure}[b]
    \includegraphics[width=1\linewidth]{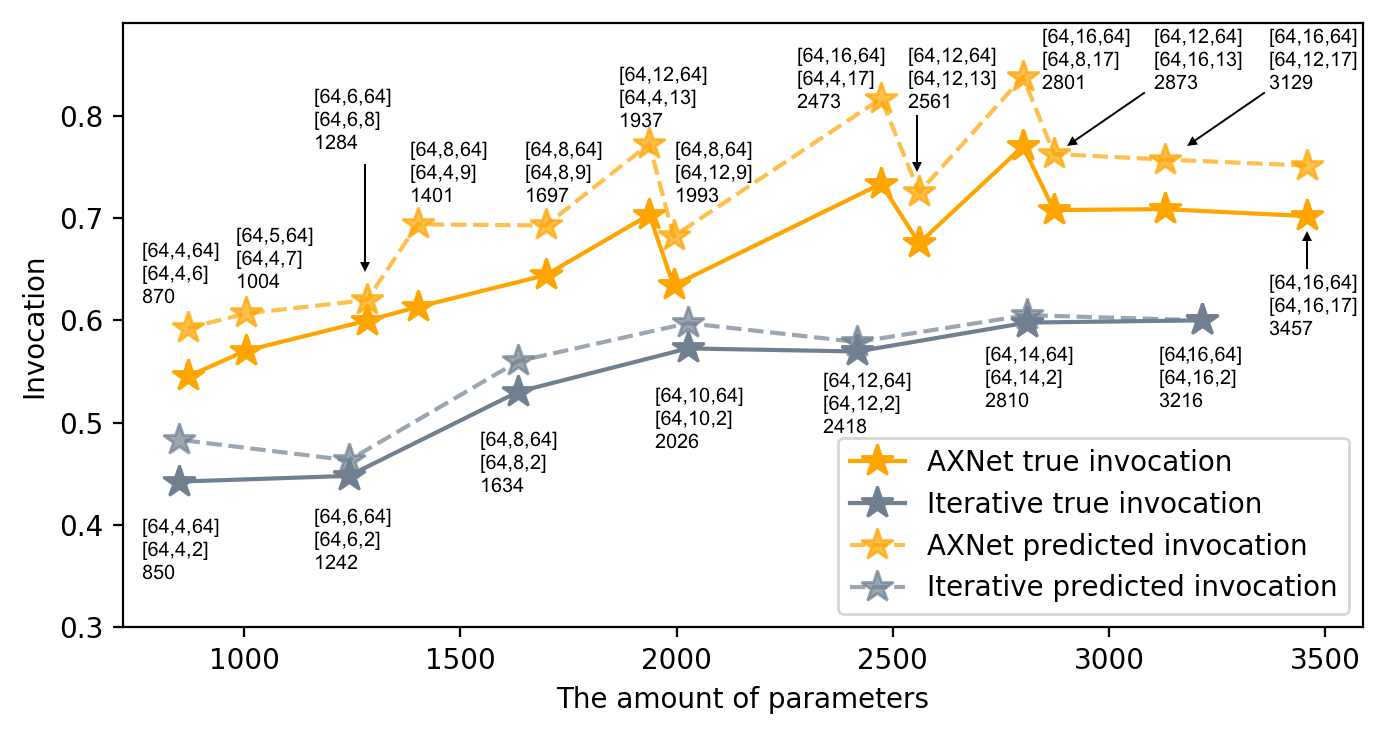}
    \caption{Comparison on the invocation in jpeg varying the parameter count. The topology of the approximation and prediction subnets, and the parameter count is labeled near the stars.}
    \label{fig:jpeg}
\end{figure}
\begin{figure*}[!tbh]
\subfigure[]{
\includegraphics[width=0.33\linewidth]{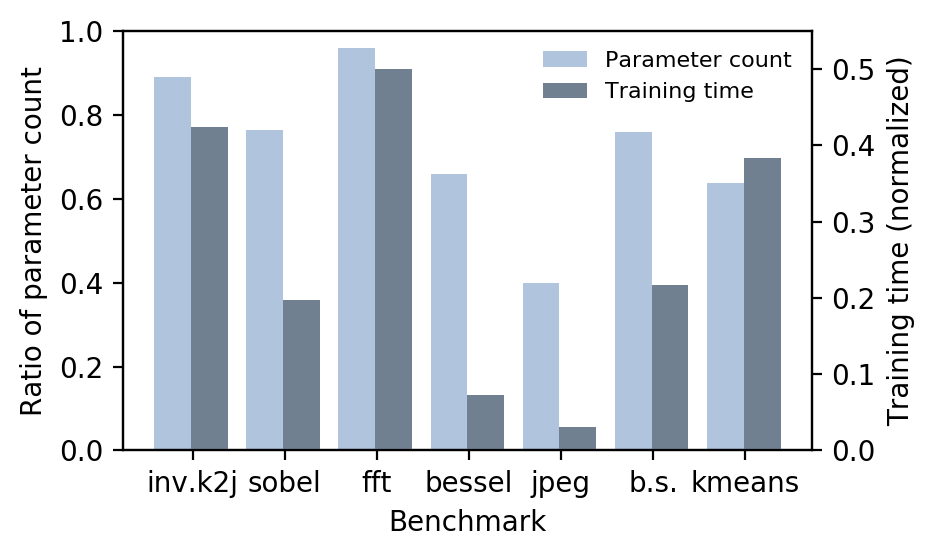}
\label{fig:ratio}
}
\subfigure[]{
    \includegraphics[width=0.3\linewidth]{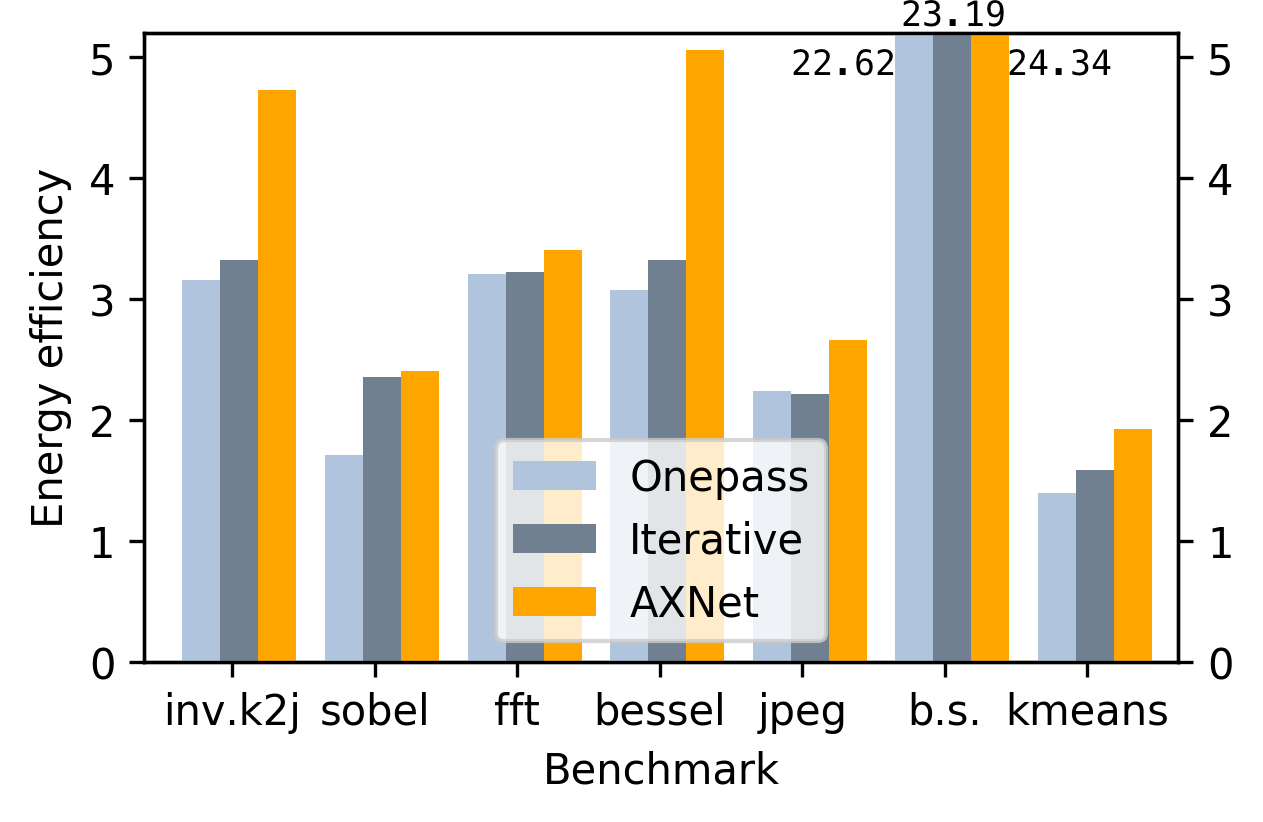}
\label{fig:energy_efficiency}}
\subfigure[]{
        \includegraphics[width=0.31\linewidth]{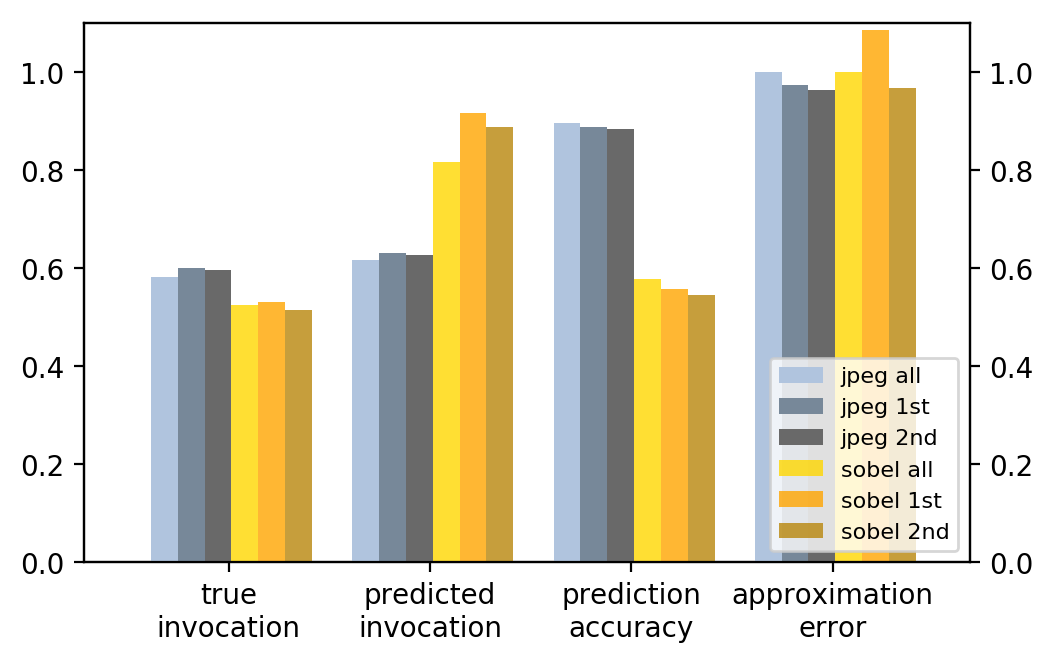}
        \label{fig:scale}}
        \vspace{-10pt}
\caption{(a) The ratio of parameter count and the training time of AXNet compared to iterative training. (b) Comparisons on the energy efficiency. (c) Investigation of the scalability of AXNet by fusing two subnets in three ways.}
\label{fig:three_figures}
\end{figure*}

Figure~\ref{fig:jpeg} illustrates the variation of the true and predicted invocation by varying the network topology, i.e., adjusting the number of neurons in hidden layers. AXNet always achieves better (true and predicted) invocation than the iterative method. When two methods achieve the same true invocation, e.g., near $60\%$, the iterative method uses a three-layer MLP with 64-16-64 neurons (namely 64-dimensional input, 16 neurons in the first hidden layer and 64-dimensional output, similarly from now on) for the approximator and an MLP with 64-16-2 neurons for the classifier. The total number of synaptic parameters is 3216. While AXNet only requires 64-6-64 for approximation subnet and 64-4-8 for prediction subnet, totally 1284 synaptic parameters. These results imply the fusion of the approximator and predictor in AXNet can eliminate tremendous redundant parameters that have little contribution to the model's performance. However, we observe that the iterative method yields more stable invocation as the neural networks becomes larger because iterative method incurs much more training effort. 

To validate the above observation in other benchmark functions, in the left side bars in Figure \ref{fig:ratio}, we demonstrate the ratio of parameter count in AXNet to that of predecessor methods when they have similar true invocation as mentioned in Section 5.1. We observe that larger structure can achieve better parameter reduction. Jpeg benchmark requires thousands of parameters and AXNet makes $60\%$ reduction of parameter count (left side bar). We also normalize the training time of AXNet to that of the iterative method. The reduction of training time is as high as 90\% in Jpeg and 74\% in average among these benchmark functions. The  right side bar in \ref{fig:ratio} shows the training time. AXNet consumes much less training time than iterative method. In the iterative method, some of the training data is intentionally discarded. The exact training times is unclear. Compared to iterative training, AXNet can achieve 13.8$times$ and 32$\times$ speedup in training time for Bessel and Jpeg. Statistics suggest that FFT incurs the least training time, and thus the reduction of training time is only 50\%. 

Figure \ref{fig:energy_efficiency} depicts the energy-efficiency. AXNet outperforms the two previous works in all benchmark applications. The cost of proposed NPU is almost identical to that in the onepass method, except the approximation subnet unit in each PE. Thus, the enhancement of the true invocation contributes to the improvement of the energy-efficiency. 

Figure \ref{fig:scale} shows the examination of the subnet fusion technique described in Section 3.4. We try two ways for the fusion of subnets: apply Hadamard product only in the first hidden layer of the approximation subnet, and only in the second hidden layer, respectively. We test their true invocation in two representative benchmark functions as they require large approximation subnet, e.g., jpeg and sobel. In jpeg, we use a AXNet with topology 64-8-8-64 as approximation subnet and 64-12-18 (connect all, 1014 parameters) or 64-12-10 (connect one hidden layer, 910 parameters) as prediction subnet. Same in sobel: 9-6-6-1 as approximation subnet and 9-8-14 (206 parameters) or 9-8-8 (162 parameters) as prediction subnet. Figure \ref{fig:scale} compares three ways of applying Hadamard product: at all hidden layers of approximation subnet~(``all''), at the first hidden layer~(``1st''), and at the second hidden layer~(``2nd''). The results suggest that these three ways of fusion make no evident difference on the performance of the AXNet, which validates the effectiveness of the subnet fusion with a single control vector.
\section {Conclusion}
\label{sect:conclusion}
This paper presents AXNet, an end-to-end trainable neural network for approximate computing with quality control. Guided by the multitask learning principle, AXNet fuses the approximator and predictor through Hadamard product. Experimental results show its superior invocation and gain of energy-efficiency over the existing neural approximate computing frameworks. We also provide the theoretical interpretation and experimental validation for AXNet's advantage in approximation error. At last, AXNet incurs much less training time and smaller scale than the existing works. In future work, we will study the compression technique for AXNet and interpret the underlying mechanism that enables AXNet. We will also evaluate the speedup and energy reduction in a real AXNet NPU implementation.
\footnotesize
\bibliographystyle{IEEEtranS.bst}
\bibliography{IEEEabrv,cite.bib}

\begin{thebibliography}{10}
\providecommand{\url}[1]{#1}
\csname url@samestyle\endcsname
\providecommand{\newblock}{\relax}
\providecommand{\bibinfo}[2]{#2}
\providecommand{\BIBentrySTDinterwordspacing}{\spaceskip=0pt\relax}
\providecommand{\BIBentryALTinterwordstretchfactor}{4}
\providecommand{\BIBentryALTinterwordspacing}{\spaceskip=\fontdimen2\font plus
\BIBentryALTinterwordstretchfactor\fontdimen3\font minus
  \fontdimen4\font\relax}
\providecommand{\BIBforeignlanguage}[2]{{%
\expandafter\ifx\csname l@#1\endcsname\relax
\typeout{** WARNING: IEEEtranS.bst: No hyphenation pattern has been}%
\typeout{** loaded for the language `#1'. Using the pattern for}%
\typeout{** the default language instead.}%
\else
\language=\csname l@#1\endcsname
\fi
#2}}
\providecommand{\BIBdecl}{\relax}
\BIBdecl

\bibitem{baxter1995learning}
J.~Baxter, ``Learning internal representations,'' pp. 311--320, 1995.

\bibitem{caruana1998multitask}
R.~Caruana, ``Multitask learning,'' in \emph{Learning to learn}.\hskip 1em plus
  0.5em minus 0.4em\relax Springer, 1998, pp. 95--133.

\bibitem{caruana1993multitask}
R.~A. Caruana, ``Multitask connectionist learning,'' in \emph{In Proceedings of
  the 1993 Connectionist Models Summer School}.\hskip 1em plus 0.5em minus
  0.4em\relax Citeseer, 1993.

\bibitem{Chen2014DaDianNao}
Y.~Chen, T.~Luo, S.~Liu, S.~Zhang, L.~He, J.~Wang, L.~Li, T.~Chen, Z.~Xu,
  N.~Sun \emph{et~al.}, ``Dadiannao: A machine-learning supercomputer,'' in
  \emph{Proceedings of the 47th Annual IEEE/ACM International Symposium on
  Microarchitecture}.\hskip 1em plus 0.5em minus 0.4em\relax IEEE Computer
  Society, 2014, pp. 609--622.

\bibitem{ernst2003razor}
D.~Ernst, N.~S. Kim, S.~Das, S.~Pant, R.~Rao, T.~Pham, C.~Ziesler, D.~Blaauw,
  T.~Austin, K.~Flautner \emph{et~al.}, ``Razor: A low-power pipeline based on
  circuit-level timing speculation,'' in \emph{Proceedings of the 36th annual
  IEEE/ACM International Symposium on Microarchitecture}.\hskip 1em plus 0.5em
  minus 0.4em\relax IEEE Computer Society, 2003, p.~7.

\bibitem{hadi2012neural}
H.~Esmaeilzadeh, A.~Sampson, L.~Ceze, and D.~Burger, ``Neural acceleration for
  general-purpose approximate programs,'' in \emph{Proceedings of the 2012 45th
  Annual IEEE/ACM International Symposium on Microarchitecture}.\hskip 1em plus
  0.5em minus 0.4em\relax IEEE Computer Society, 2012, pp. 449--460.

\bibitem{2016GAN}
I.~Goodfellow, ``Nips 2016 tutorial: Generative adversarial networks,''
  \emph{arXiv preprint arXiv:1701.00160}, 2016.

\bibitem{goodfellow2016deep}
I.~Goodfellow, Y.~Bengio, A.~Courville, and Y.~Bengio, \emph{Deep
  learning}.\hskip 1em plus 0.5em minus 0.4em\relax MIT press Cambridge, 2016,
  vol.~1.

\bibitem{han2016eie}
S.~Han, X.~Liu, H.~Mao, J.~Pu, A.~Pedram, M.~Horowitz, and W.~J. Dally, ``Eie:
  efficient inference engine on compressed deep neural network,''
  \emph{international symposium on computer architecture}, vol.~44, no.~3, pp.
  243--254, 2016.

\bibitem{He2016ACR}
X.~He, G.~Yan, Y.~Han, and X.~Li, ``Acr: Enabling computation reuse for
  approximate computing,'' in \emph{Design Automation Conference}, 2016, pp.
  643--648.

\bibitem{Hornik1991Approximation}
K.~Hornik, ``Approximation capabilities of multilayer feedforward networks,''
  \emph{Neural networks}, vol.~4, no.~2, pp. 251--257, 1991.

\bibitem{khudia2015rumba}
D.~S. Khudia, B.~Zamirai, M.~Samadi, and S.~Mahlke, ``Rumba: An online quality
  management system for approximate computing,'' in \emph{Computer Architecture
  (ISCA), 2015 ACM/IEEE 42nd Annual International Symposium on}.\hskip 1em plus
  0.5em minus 0.4em\relax IEEE, 2015, pp. 554--566.

\bibitem{lawrence1998neural}
S.~Lawrence, I.~Burns, A.~D. Back, A.~C. Tsoi, and C.~L. Giles, ``Neural
  network classification and prior class probabilities,'' \emph{neural
  information processing systems}, pp. 299--313, 1998.

\bibitem{longadge2013class}
R.~Longadge and S.~Dongre, ``Class imbalance problem in data mining review,''
  \emph{arXiv preprint arXiv:1305.1707}, 2013.

\bibitem{Mahajan2015neuralprediction}
D.~Mahajan, A.~Yazdanbakhsh, J.~Park, B.~Thwaites, and H.~Esmaeilzadeh,
  ``Prediction-based quality control for approximate accelerators,'' in
  \emph{Second Workshop on Approximate Computing Across the System Stack,
  WACAS}, 2015.

\bibitem{Mahajan2016Towards}
------, ``Towards statistical guarantees in controlling quality tradeoffs for
  approximate acceleration,'' \emph{Acm Sigarch Computer Architecture News},
  vol.~44, no.~3, pp. 66--77, 2016.

\bibitem{samadi2013sage:}
M.~Samadi and et~al., ``Sage: self-tuning approximation for graphics engines,''
  \emph{international symposium on microarchitecture}, pp. 13--24, 2013.

\bibitem{sui2016proactive}
X.~Sui, A.~Lenharth, D.~S. Fussell, and K.~Pingali, ``Proactive control of
  approximate programs,'' \emph{ACM SIGOPS Operating Systems Review}, vol.~50,
  no.~2, pp. 607--621, 2016.

\bibitem{wang2016effective}
T.~Wang, Q.~Zhang, N.~S. Kim, and Q.~Xu, ``On effective and efficient quality
  management for approximate computing,'' in \emph{Proceedings of the 2016
  International Symposium on Low Power Electronics and Design}.\hskip 1em plus
  0.5em minus 0.4em\relax ACM, 2016, pp. 156--161.

\bibitem{Xu2017Iterative}
C.~Xu, X.~Wu, W.~Yin, Q.~Xu, N.~Jing, X.~Liang, and L.~Jiang, ``On quality
  trade-off control for approximate computing using iterative training,'' in
  \emph{Proceedings of the 54th Annual Design Automation Conference
  2017}.\hskip 1em plus 0.5em minus 0.4em\relax ACM, 2017, p.~52.

\bibitem{xu2015exploring}
X.~Xu and H.~H. Huang, ``Exploring data-level error tolerance in
  high-performance solid-state drives,'' \emph{IEEE Transactions on
  Reliability}, vol.~64, no.~1, pp. 15--30, 2015.

\bibitem{2016axbench}
A.~Yazdanbakhsh, D.~Mahajan, H.~Esmaeilzadeh, and P.~Lotfi-Kamran, ``Axbench: A
  multiplatform benchmark suite for approximate computing,'' \emph{IEEE Design
  \& Test}, vol.~34, no.~2, pp. 60--68, 2017.

\bibitem{zabokrtsky2015feature}
Z.~Zabokrtsky, ``Feature engineering in machine learning,'' \emph{Institute of
  Formal and Applied Linguistics, Charles University in Prague}, 2015.

\bibitem{zhang2015approxann:}
Q.~Zhang and et~al., ``Approxann: an approximate computing framework for
  artificial neural network,'' in \emph{Proceedings of the 2015 Design,
  Automation \& Test in Europe Conference \& Exhibition}.\hskip 1em plus 0.5em
  minus 0.4em\relax EDA Consortium, 2015, pp. 701--706.

\end{thebibliography}
\end{document}